\documentclass[letterpaper]{article} 
\usepackage[]{aaai25}  
\usepackage{times}  
\usepackage{helvet}  
\usepackage{courier}  
\usepackage[hyphens]{url}  
\usepackage{graphicx} 
\usepackage{natbib}  
\usepackage{caption} 
\usepackage{multirow}
\usepackage{algorithm}
\usepackage{algorithmic}
\usepackage{amsmath}
\usepackage{newfloat}
\usepackage{listings}
\DeclareCaptionStyle{ruled}{labelfont=normalfont,labelsep=colon,strut=off} 
\floatstyle{ruled}
\newfloat{listing}{tb}{lst}{}
\floatname{listing}{Listing}
\title{Training on the Benchmark Is Not All You Need}
\author{
	Shiwen Ni\textsuperscript{\rm 1,2,$^\star$}, Xiangtao Kong\textsuperscript{\rm 1,3,$^\star$}, Chengming Li\textsuperscript{\rm 4}, Xiping Hu\textsuperscript{\rm 4}, Ruifeng Xu\textsuperscript{\rm 5}\\Jia Zhu\textsuperscript{\rm 2,}$^\ast$, Min Yang \textsuperscript{\rm 1,2,}\thanks{Corresponding author}
}
\affiliations{
	\textsuperscript{\rm 1}Shenzhen Key Laboratory for High Performance Data Mining, Shenzhen Institutes of Advanced Technology, CAS\\
    \textsuperscript{\rm 2}Key Laboratory of Intelligent Education Technology and Application of Zhejiang Province, Zhejiang Normal University \\
	\textsuperscript{\rm 3}University of Science and Technology of China  \\
	\textsuperscript{\rm 4}Shenzhen MSU-BIT University \\
	\textsuperscript{\rm 5}Harbin Institute of Technology (Shenzhen) 
	
	\{sw.ni, min.yang\}@siat.ac.cn, ~jiazhu@zjnu.edu.cn
}

\usepackage{bibentry}

\begin{document}

\maketitle

\begin{abstract}
The success of Large Language Models (LLMs) relies heavily on the huge amount of pre-training data learned in the pre-training phase. The opacity of the pre-training process and the training data causes the results of many benchmark tests to become unreliable. If any model has been trained on a benchmark test set, it can seriously hinder the health of the field. In order to automate and efficiently test the capabilities of large language models, numerous mainstream benchmarks adopt a multiple-choice format. As the swapping of the contents of multiple-choice options does not affect the meaning of the question itself, we propose a simple and effective data leakage detection method based on this property. Specifically, we shuffle the contents of the options in the data to generate the corresponding derived data sets, and then detect data leakage based on the model's log probability distribution over the derived data sets. If there is a maximum and outlier in the set of log probabilities, it indicates that the data is leaked. Our method is able to work under gray-box conditions without access to model training data or weights, effectively identifying data leakage from benchmark test sets in model pre-training data, including both normal scenarios and complex scenarios where options may have been shuffled intentionally or unintentionally. Through experiments based on two LLMs and benchmark designs, we demonstrate the effectiveness of our method. In addition, we evaluate the degree of data leakage of 35 mainstream open-source LLMs on four benchmark datasets and give a ranking of the leaked LLMs for each benchmark, and we find that the Qwen family of LLMs has the highest degree of data leakage.
\end{abstract}

\begin{links}
\link{Code}{https://github.com/nishiwen1214/Benchmark-leakage-detection}
\end{links}
%

\section{Introduction}
\begin{figure}[t]
	\centering
	\includegraphics[width=0.47\textwidth]{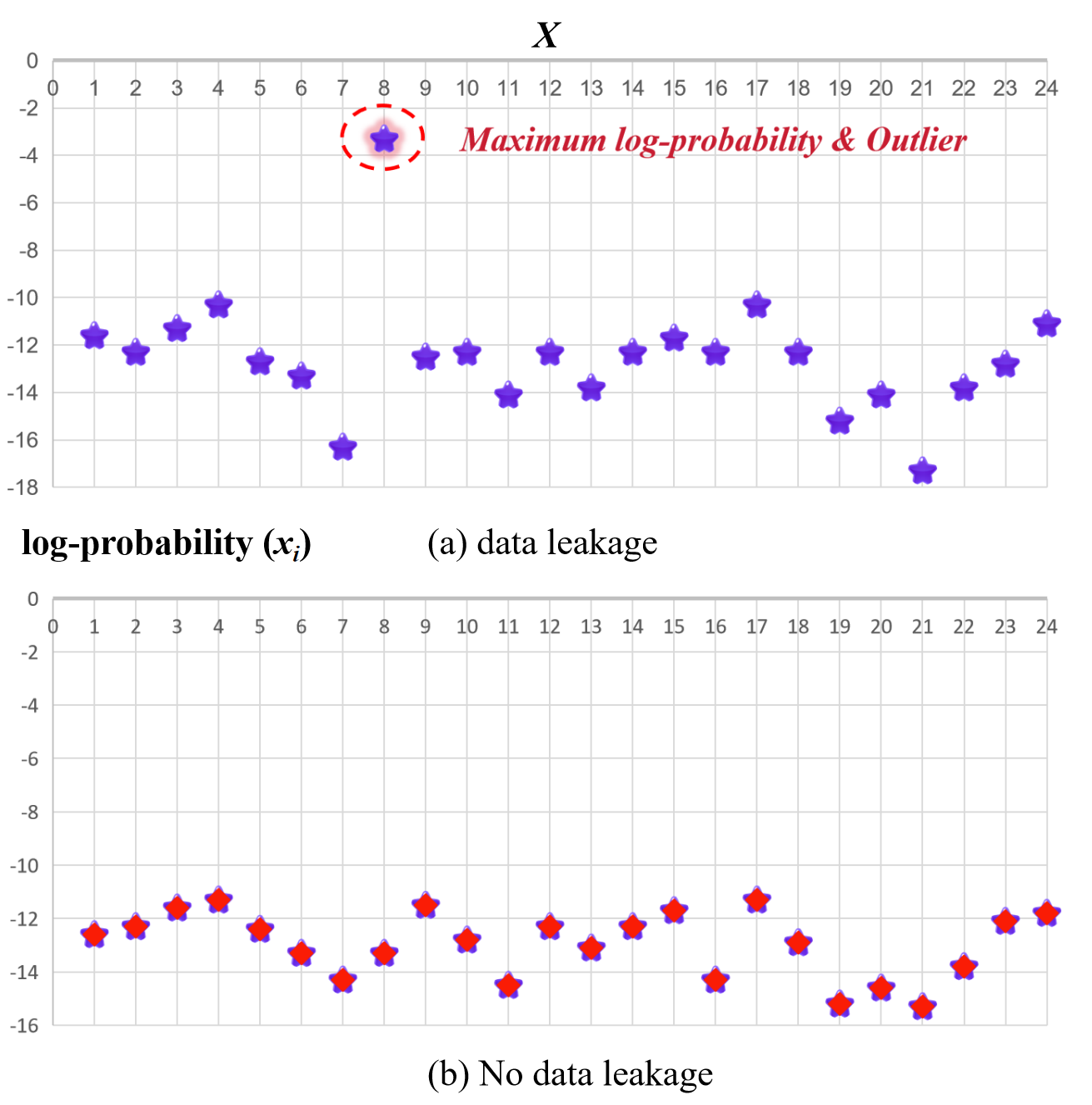} 
	\caption{Log-probability distributions for different option orders. For example: \textit{\{\textbf{Order1}: All of the following are examples of connective tissue EXCEPT A: \underline{ligaments} B: \underline{muscle} C: \underline{blood} D: \underline{cartilage} ,..., \textbf{Order24}: All of the following are examples of connective tissue EXCEPT A: \underline{cartilage} B: \underline{blood} C: \underline{muscle} D: \underline{ligaments}\}}.
	}
	\label{fig1}
\end{figure}
Recently, large language models (LLMs) have made significant advances in most natural language processing benchmarks \cite{hendrycks2021measuring,li2023cmmlu,huang2024c,wang2023cmb,cobbe2021GSM8K,zheng2024judging}. One of the key reasons why LLMs have achieved such success is through large-scale pre-training on large corpora collected from the Internet. However, due to the intentional or unintentional data collection process of the developers of LLMs, the pre-trained corpus may set contain data from various evaluation benchmarks. Data leakage from such benchmarks causes an inability to accurately evaluate the true performance of LLMs, and the model may simply memorize the answers to difficult questions. The composition of the pre-trained corpus is often considered to be the core secret of existing large models, and open-source models such as LLaMA \cite{touvron2023llama}, Qwen \cite{bai2023qwen}, and Yi \cite{young2024yi} do not open-source the full training data of their models.
Currently most LLMs do not disclose their full pre-training data, which makes it uncertain whether the performance of these LLMs on certain benchmarks is realistic and credible.
There is growing concern about the proper use of benchmarks and fair comparisons between different models. \citet{zhou2023don} investigated the impact of benchmark leakage and found that when the pre-training data of a large language model includes data from one of the review benchmarks, it will perform better in this evaluation benchmark, but its performance will drop in other irrelevant tasks, ultimately leading to unreliable assessments of the model's performance.

Many companies and research organizations often advertise how many scores their LLMs have achieved on various benchmarks, achieving first place, yet the fairness of that score is not taken seriously.
Some of the current mainstream benchmarks (e.g., MMLU \cite{hendrycks2021measuring}, CMMLU \cite{li2023cmmlu}, C-Eval \cite{huang2024c}, E-Eval \cite{hou2024eval}, CMB \cite{wang2023cmb}) are in the form of multiple-choice questions. Theoretically, by changing the order of the content of the options, the model predicts that the logarithmic probability of that data may become higher or lower, but the fluctuation will not be very large. For example, if the model has not been trained on either order of data, the log probabilities of “All of the following are examples of connective tissue EXCEPT A: \underline{ligaments} B: \underline{muscle} C: \underline{blood} D: \underline{cartilage}” and “All of the following are examples of connective tissue EXCEPT A: \underline{cartilage} B: \underline{blood} C: \underline{muscle} D: \underline{ligaments}” will not differ much because of the lack of sequential relationship between the contents of the options.
As shown in Figure 1, a data containing four options can be composed into 4!=24 different derived data after shuffling the contents of the options. Without knowing the order of the options in the pre-training data (the order of the shuffled options may be assumed during the benchmark construction process or the pre-training data construction process \cite{huang2024c,hou2024eval}), as in Fig. 1(b), if the 24 log probabilities are both high and low without a very large value of some kind, then there is no data leakage; if there is a significant outlier with the maximum of the log probabilities, as shown in Fig. (a), then there is a data leakage. With this detection method, artificial and intentional shuffling over the order of options can also be detected, if the option shuffling is not taken into account, only the logarithmic probability of the data in the original order is required to maximize the probability of data leakage can be determined.

In this work, we show how to provide reliable evidence for test set contamination in gray-box language models (Only need to access the Log-Probability of the output). More specifically, we provide a simple and efficient new method for benchmark leakage detection based on multiple-choice questions. The method identifies the presence of a benchmark test set in a language model's pre-training data and the extent of data leakage without accessing the model's training data or weights. 
The contributions of this paper are summarized below:

\begin{itemize}
    \item We propose a simple yet effective detection method based on the characteristics of multiple choice questions by generating different derived datasets by disrupting the order of the options, and then using the model's logarithmic probability distribution to detect whether the original dataset is leaked or not.
    \item The algorithms are able to work in gray-box conditions without access to model training data or weights, effectively identifying data leakage from the benchmark test set in the model pre-training data, including normal scenarios and complex scenarios in which options may have been intentionally or unintentionally disrupted.
    \item We validate the effectiveness of the approach based on two LLMs design experiments and evaluate the data leakage risk of 35 open-source LLMs on four mainstream benchmark sets, present a benchmark leakage leaderboard among LLMs, and in particular find that the Qwen family of LLMs shows a high risk in several benchmarks.
\end{itemize}

\section{Related Work}
\subsection{Mainstream Benchmarks for LLMs}
As natural language processing enters the LLM era, a wide variety of LLMs \citep{team2023internlm,touvron2023llama,touvron2023llama2,young2024yi,BAAI,bai2023qwen,yang2023baichuan} have emerged. Various comprehensive or specialized benchmarks \cite{zhong2023agieval,zheng2024judging,cobbe2021GSM8K,hendrycks2021math,hendrycks2021measuring,li2023cmmlu,huang2024c,wang2023cmb} have also been proposed to accurately assess various aspects of the model's capabilities. In order to automate and efficiently test the capabilities of large language models, many mainstream benchmarks use a multiple-choice format. For example, MMLU \cite{hendrycks2021measuring} is a comprehensive and all-encompassing English benchmark, CMMLU \cite{li2023cmmlu} and C-Eval \cite{huang2024c} are comprehensive and all-encompassing Chinese benchmarks, and CMB \cite{wang2023cmb} is a comprehensive and all-encompassing Chinese medical quiz assessment benchmark. In addition, multimodal comprehension benchmarks such as MMMU \cite{yue2024mmmu} and CMMMU \cite{zhang2024cmmmu} are also in the form of multiple choice questions. This work focuses on the problem of benchmark test set leakage in the form of multiple-choice questions. Since the exchange of multiple-choice question option content does not affect the meaning of the question itself, we propose a simple and effective data leakage detection method based on this property.
\subsection{Data Leakage Detection}
The current pre-training model size and its pre-training corpus are getting larger and larger, which inevitably leads to data leakage between the pre-training corpus and various benchmark test sets. Several previous studies \cite{brown2020language,wei2021finetuned} have utilized post-hoc n-gram overlap analysis between the benchmark and pre-training corpus to measure data leakage. \citet{deng2023investigating} utilized benchmark perturbations and synthetic data to detect benchmark leakage. \cite{wei2023skywork} compare the model's loss on the training, validation, and test sets; if the model's loss on the training set is significantly lower than on the validation or test sets, this may indicate that the model is overfitting the training data. If the loss on the test set is significantly lower than an independent reference set (consisting of data that the model has never seen), this may indicate that the test data was compromised during training. \citet{mattern2023membership} tested the difference in perplexity between the target sequence and the randomized sequence. \citet{oren2023proving} exchanged the order of problems in some benchmarks and tested the model with generating new data as a way to detect data leakage. \citet{xu2024benchmarking} introduced a measure of the predictive accuracy of the benchmark model using two simple and scalable metrics, complexity and n-gram accuracy, to identify potential data leakage. \citet{dong-etal-2024-generalization} identified data contamination by analysing the peakedness of the model output distribution. Our work leverages the interchangeability of options in benchmark test sets to achieve instance and fine-grained data leakage detection. In addition to common scenarios, we even consider data leakage identification in scenarios where options are intentionally or unintentionally shuffled.
\begin{figure*}[t]
	\centering
	\includegraphics[width=1\textwidth]{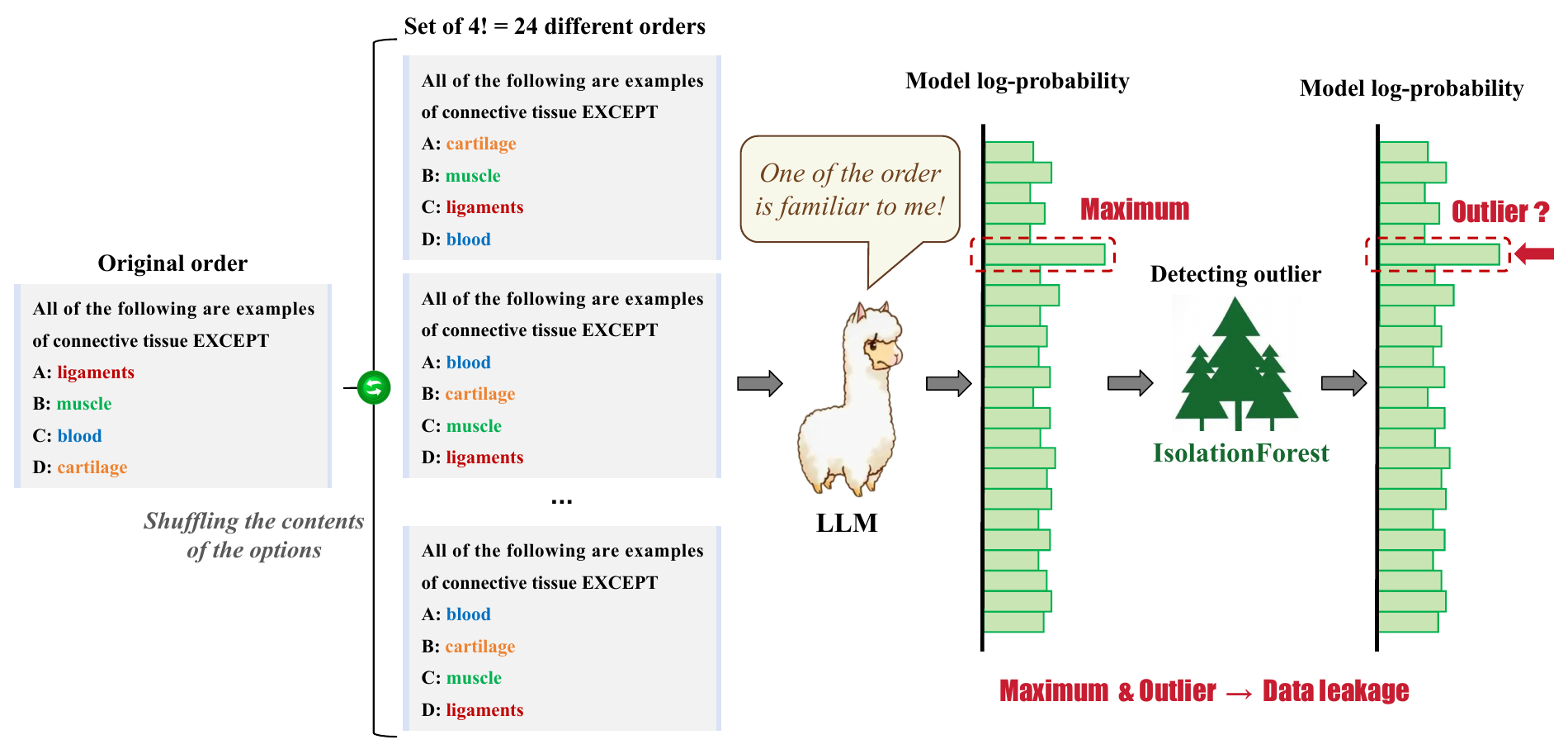} 
	\caption{The order with the largest probability value, which is an outlier, indicates that the data in that order was pre-trained.}
	\label{fig2}
\end{figure*}

\begin{algorithm}[tb]
	\caption{Data Leakage Detection Under Scenario (\textit{a})}
	\label{alg1}
	\textbf{Input}: 
	\begin{itemize}  
		\item Data to be detected: $x=[q,o_1,o_2,...,o_n]$  
		\item Target Model: $\mathcal{M}$  
	\end{itemize} 
	\textbf{Output}: Whether the data was leaked (“L” for Leaked, “NL” for Not Leaked)
	\begin{algorithmic}[1] 
		\STATE Get the set of n! derived data $\mathcal{X}$:
		\[
		\mathrm{Shuffle}(x)\rightarrow \mathcal{X} = \{x^*_1,x_2,...,x_{n!}\} 
		\]
		\FOR{each derived data $x_i$}
		\STATE Calculate the log probability of the derived data:
		\[
		\mathrm{log}p_i = P^\mathcal{M}(\mathrm{seq}[q,\mathrm{Shuffle}^i(o_1,o_2,...,o_n)])
		\]
		\ENDFOR 
		\STATE Get the set of n! log probabilities $\mathrm{log}p_i$:
		\[
		\mathcal{P}=\{\mathrm{log}p^*_1,\mathrm{log}p_2,...,\mathrm{log}p_n\}
		\]
		\IF {$\mathrm{log}p^*_1=\mathrm{max}(\mathcal{P})$}
		\STATE \textbf{return} “L”
		\ELSE
		\STATE \textbf{return} “NL”
		\ENDIF
	\end{algorithmic}
\end{algorithm} 

\begin{algorithm}[tb]
	\caption{Data Leakage Detection Under Scenario (\textit{b})}
	\label{alg2}
	\textbf{Input}: 
	\begin{itemize}  
		\item Data to be detected: $x=[q,o_1,o_2,...,o_n]$  
		\item Target Model: $\mathcal{M}$  
		\item Outlier threshold: $\delta$
	\end{itemize} 
	\textbf{Output}: Whether the data was leaked (“L” for Leaked, “NL” for Not Leaked)
	\begin{algorithmic}[1] 
		\STATE Get the set of n! derived data $\mathcal{X}$:
		\[
		\mathrm{Shuffle}(x)\rightarrow \mathcal{X} = \{x_1,x_2,...,x_{n!}\} 
		\]
		\FOR{each derived data $x_i$}
		\STATE Calculate the log probability of the derived data:
		\[
		\mathrm{log}p_i = P^\mathcal{M}(\mathrm{seq}[q,\mathrm{Shuffle}^i(o_1,o_2,...,o_n)])
		\]
		\ENDFOR 
		\STATE Get the set of n! log probabilities $\mathrm{log}p_i$:
		\[
		\mathcal{P}=\{\mathrm{log}p_1,\mathrm{log}p_2,...,\mathrm{log}p_{n!}\}
		\]
		\STATE Calculate the outlier score $s^{out}_i$ for each data:
		\[
		\mathcal{S}^{out}=\{s^{out}_1,s^{out}_2,...,s^{out}_{n!}\} \leftarrow \mathrm{IsolationForest}(\mathcal{P})
		\]
		\STATE Obtain the maximum log probability $\mathrm{log}p_m$ and corresponding outlier score $s^{out}_m$:
		\[
		s^{out}_m\leftarrow \mathrm{log}p_m \leftarrow \mathrm{max}(\mathcal{P})
		\]
		\IF {$s^{out}_m<\delta$}
		\STATE \textbf{return} “L”
		\ELSE
		\STATE \textbf{return} “NL”  
		\ENDIF
	\end{algorithmic}
\end{algorithm} 
\section{Methodology}
Our goal is to identify whether the pre-training process of a language model $\theta$ includes a particular piece of data $x$ from a benchmark test set, or the extent to which that benchmark test set $D$ leaks to the model $\theta$. Detection in our setup is under gray-box conditions, i.e., the pre-training corpus and parameters of the model are unknown. We consider two scenarios: (\textit{a}) where the order in which the pre-trained data options are presented is not shuffled, and (\textit{b}) where the sequence of pre-trained data options may be shuffled.
\subsection{Scenario \textit{a}: Not Shuffled}
As illustrated in Algorithm \ref{alg1}, we present the pseudo-code for a data leakage detection method under the scenario where the options are not shuffled. We define a piece of data to be tested as \( x = [q, o_1, o_2, \ldots, o_n] \), where \( q \) is the question in a multiple-choice format, \( o_i \) is the \( i \)-th option, and \( n \) is the total number of options.

As depicted in Figure 2, subjecting the data \( x \) to an option shuffle operation yields a derived dataset \( \mathcal{X} \), expressed as \( \mathrm{Shuffle}(x) \rightarrow \mathcal{X} = \{x^*_1, x_2, \ldots, x_{n!} \} \). Here, \( \mathrm{Shuffle} \) denotes the function for shuffling options, capable of generating \( n! \) distinct permutations, with \( n \) representing the number of options.

When considering the possibility that the options within the data have not been artificially rearranged, \( x^*_1 \) is identified as the original data sequence. Subsequently, each \( x_i \in \mathcal{X} \) is fed into the target model \( \mathcal{M} \) to calculate the respective log probability, denoted by:
\[ \mathrm{log}p_i = P^{\mathcal{M}}(\mathrm{seq}[q, \mathrm{Shuffle}^i(o_1, o_2, \ldots, o_n)]) \]
These probabilities are then compiled into the set \( \mathcal{P} = \{\mathrm{log}p_1, \mathrm{log}p_2, \ldots, \mathrm{log}p_{n!} \} \), where \( \mathrm{log}p_1 \) corresponds to the original sequence \( x^*_1 \).

The detection criterion is based on the comparison of \( \mathrm{log}p_1 \) against the values within \( \mathcal{P} \). If \( \mathrm{log}p_1 \) is the maximum value within \( \mathcal{P} \), this suggests that the data has been influenced by the training of the model \( \mathcal{M} \), and we conclude that the data has leaked.

\subsection{Scenario \textit{b}: Shuffled} 
The pseudo-code of the data leakage detection method under scenario \textit{b} is presented in Algorithm \ref{alg2}. Under these conditions, the data in the test set can be shuffled through, and any kind of sequence order may be the order fitted by the model. As above, we first shuffle the data to be tested to get n! derived data: $\mathrm{Shuffle}(x)\rightarrow \mathcal{X} = \{x_1,x_2,...,x_{n!}\}$. Then, we process each derived data point \( x_i \). Specifically, we calculate the log probability of the derived data using the following formula:
\[
\mathrm{log}p_i = P^{\mathcal{M}}(\mathrm{seq}[q, \mathrm{Shuffle}^i(o_1, o_2, \ldots, o_n)])
\]
Here, \( P^{\mathcal{M}} \) represents the probability distribution under model \( \mathcal{M} \), \( \mathrm{seq} \) denotes the sequence, \( q \) is the question, \( \mathrm{Shuffle}^i \) is the \( i \)-th shuffle operation, and \( o_1, o_2, \ldots, o_n \) are the original data points. As depicted in Figure 2, we calculate this for all possible shuffle combinations, obtaining a set \( \mathcal{P} \) of \( n! \) log probabilities:
\[
\mathcal{P} = \{\mathrm{log}p_1, \mathrm{log}p_2, \ldots, \mathrm{log}p_{n!}\}
\]
Next, we calculate the outlier score \( s^{out}_i \) for each data point using an isolation forest algorithm:
\[
\mathcal{S}^{out} = \{s^{out}_1, s^{out}_2, \ldots, s^{out}_{n!}\} \leftarrow \mathrm{IsolationForest}(\mathcal{P})
\]
Subsequently, we identify the maximum log probability \( \mathrm{log}p_m \) and its corresponding outlier score \( s^{out}_m \) by taking the maximum value from the set \( \mathcal{P} \):
\[
s^{out}_m \leftarrow \mathrm{log}p_m \leftarrow \mathrm{max}(\mathcal{P})
\]
We then evaluate whether the outlier score \( s^{out}_m \) is below a predefined threshold \( \delta \). If \( s^{out}_m < \delta \), the data is classified as an outlier ("L") and the algorithm returns this label. Otherwise, it is classified as non-outlier ("NL") and the algorithm returns this label accordingly:
\[
\begin{cases}
	\text{If } s^{out}_m < \delta: & \textbf{return} \text{"L"} \\
	\text{Otherwise}: & \textbf{return} \text{"NL"}
\end{cases}
\]

\begin{table}[]
	\centering
	\begin{tabular}{ccccc}
		\hline
		\multicolumn{5}{c}{MMLU-LLaMA2}                \\ \hline
		Epoch & Accuracy & Precision & Recall & F1-score     \\ \hline
		1     & 0.710    & 0.760     & 0.620  & 0.680  \\
		2     & 0.790    & 0.808     & 0.760  & 0.783  \\
		3     & 0.875    & 0.835     & 0.934  & 0.881  \\
		5     & 0.886    & 0.843     & 0.948  & 0.892  \\
		10    & 0.909    & 0.863     & 0.972  & 0.914  \\ \hline \hline
		\multicolumn{5}{c}{CMMLU-Qwen2}                \\ \hline
		Epoch & Accuracy & Precision & Recall & F1-score    \\ \hline
		1     & 0.603    & 0.824     & 0.262  & 0.397  \\
		2     & 0.745    & 0.918     & 0.538  & 0.678  \\
		3     & 0.888    & 0.945     & 0.824  & 0.880   \\
		5     & 0.966    & 0.955     & 0.978  & 0.966  \\
		10    & 0.974    & 0.950     & 1.000  & 0.974 \\ \hline
	\end{tabular}
	\caption{Experiment results under scenario (\textit{a}).}
	\label{tab1}
\end{table}

\begin{table*}[t]
	\centering
	\begin{tabular}{cccccccccc}
		\hline
		\multirow{2}{*}{$\delta$}      & \multirow{2}{*}{Epoch} & \multicolumn{4}{c}{MMLU-LLaMA2}       & \multicolumn{4}{c}{CMMLU-Qwen2}       \\ \cline{3-10} 
		&                        & Accuracy & Precision & Recall & F1-score    & Accuracy & Precision & Recall & F1-score    \\ \hline
		\multirow{5}{*}{-0.20}  & 1                      & 0.514    & 0.525     & 0.294  & 0.376 & 0.495    & 0.483     & 0.150  & 0.229 \\
		& 2                      & 0.570    & 0.612     & 0.380  & 0.469 & 0.548    & 0.601    & 0.284  & 0.385 \\
		& 3                      & 0.651    & 0.668     & 0.598  & 0.631 & 0.675    & 0.742     & 0.536  & 0.622 \\
		& 5                      & 0.741    & 0.728     & 0.768  & 0.747 & 0.817    & 0.794    & 0.856  & 0.823 \\
		& 10                     & 0.803    & 0.767     & 0.870  & 0.815 & 0.848    & 0.805     & 0.918  & 0.857 \\ \hline
		\multirow{5}{*}{-0.17} & 1                      & 0.517    & 0.517     & 0.498  & 0.507 & 0.508    & 0.512     & 0.340  & 0.408 \\
		& 2                      & 0.592    & 0.586     & 0.626  & 0.600 & 0.566    & 0.573     & 0.516  & 0.543 \\
		& 3                      & 0.673    & 0.631     & 0.830  & 0.717 & 0.663    & 0.646     & 0.718  & 0.680 \\
		& 5                      & 0.712    & 0.649     & 0.922  & 0.762 & 0.761    & 0.687     & 0.956  & 0.800 \\
		& 10                     & 0.734    & 0.660     & 0.962  & 0.783 & 0.763    & 0.682     & 0.984  & 0.805 \\ \hline
		\multirow{5}{*}{-0.15} & 1                      & 0.509    & 0.507     & 0.632  & 0.562 & 0.500    & 0.500     & 0.474  & 0.486 \\
		& 2                      & 0.597    & 0.571     & 0.780  & 0.659 & 0.570    & 0.557     & 0.674  & 0.610 \\
		& 3                      & 0.656    & 0.603     & 0.912  & 0.726 & 0.622    & 0.588     & 0.810  & 0.681 \\
		& 5                      & 0.666    & 0.603     & 0.970  & 0.743 & 0.699    & 0.626     & 0.984  & 0.763 \\
		& 10                     & 0.671    & 0.605     & 0.980  & 0.748 & 0.702    & 0.626     & 0.998  & 0.770 \\ \hline
	\end{tabular}
	\caption{Experiment results on MMLU and CMMLU datasets under scenario (\textit{b}).}
	\label{tab2}
\end{table*}
\section{Experiment}
\subsection{Experimental settings}
We randomly selected 1,000 pieces of data from MMLU, 500 of which were used for continuous pre-training of the LLaMA2-7b-base model, and then used these 1,000 pieces of data to test the pre-trained model, detecting which of these 1,000 pieces of data had been trained. Similarly we also used CMMLU data to test the Qwen2-7b-base model. Our experiments consider two scenarios where (\textit{a}) the order of pre-trained data options is not shuffled and (\textit{b}) the order of pre-trained data options may be shuffled.

\subsection{Experimental results}
The experimental results for scenario (\textit{a}) are shown in Table 1. Under scenario (a), as long as the log probability of each of the other 23 variants of a piece of data is smaller than that of its original order, then we predict that there is a leak in this piece of data. For LLaMA2-7B, the detection accuracy and F1 exceeded 90\% when the data were trained 10 times. We found that even if the data was only pre-trained once, our detection method was able to achieve an accuracy of 71\%, which is a passing grade. In the early stage, the accuracy of our data leakage detection increases dramatically with each increase in the number of training sessions, e.g., the accuracy reaches 79\% with an epoch of 2. For the Qwen2-7B model on the Chinese benchmark data CMMLU, the accuracy is only 60.3\% when epoch is 1, however, when epoch is 5 the accuracy is already 96.6\%. The experimental results in Table 1 show that under scenario (\textit{a}), the detection accuracy of our data leakage can achieve good performance, even with very few data duplications.

The experimental results for scenario (\textit{b}) are shown in Table \ref{tab2}. For the determination of outliers, we chose three thresholds of -0.2, 0.17, and -0.15. Since scenario \textit{b} is very challenging, the detection accuracy of scenario \textit{b} is quite lower than scenario \textit{a} from the experimental results. The highest accuracy is achieved when outlier threshold $\delta=0.2$.  When the data is trained 10 times, both accuracy and F1 on LLaMA2-7B exceed 0.8, and for Qwen2-7B even an accuracy of 84.8\% and an F1 score of 0.857 are achieved. Even if the data is only pre-trained once, our detection method is able to achieve about 50\% accuracy. From the experimental results we can choose a smaller outlier threshold when the number of training times is small. And the test results on the Chinese and English datasets are similar. However, overall the accuracy is higher on Qwen2-7B with CMMLU than on LLaMA2-7B with MMLU. We find that the recall is very low when the number of training iterations is small, and the recall improves very significantly when the number of training iterations is increased. Overall, our data leakage detection method achieves excellent accuracy in scenario \textit{a} and passable results in the challenging scenario \textit{b}.

\begin{figure*}[t]
	\centering
	\includegraphics[width=0.98\textwidth]{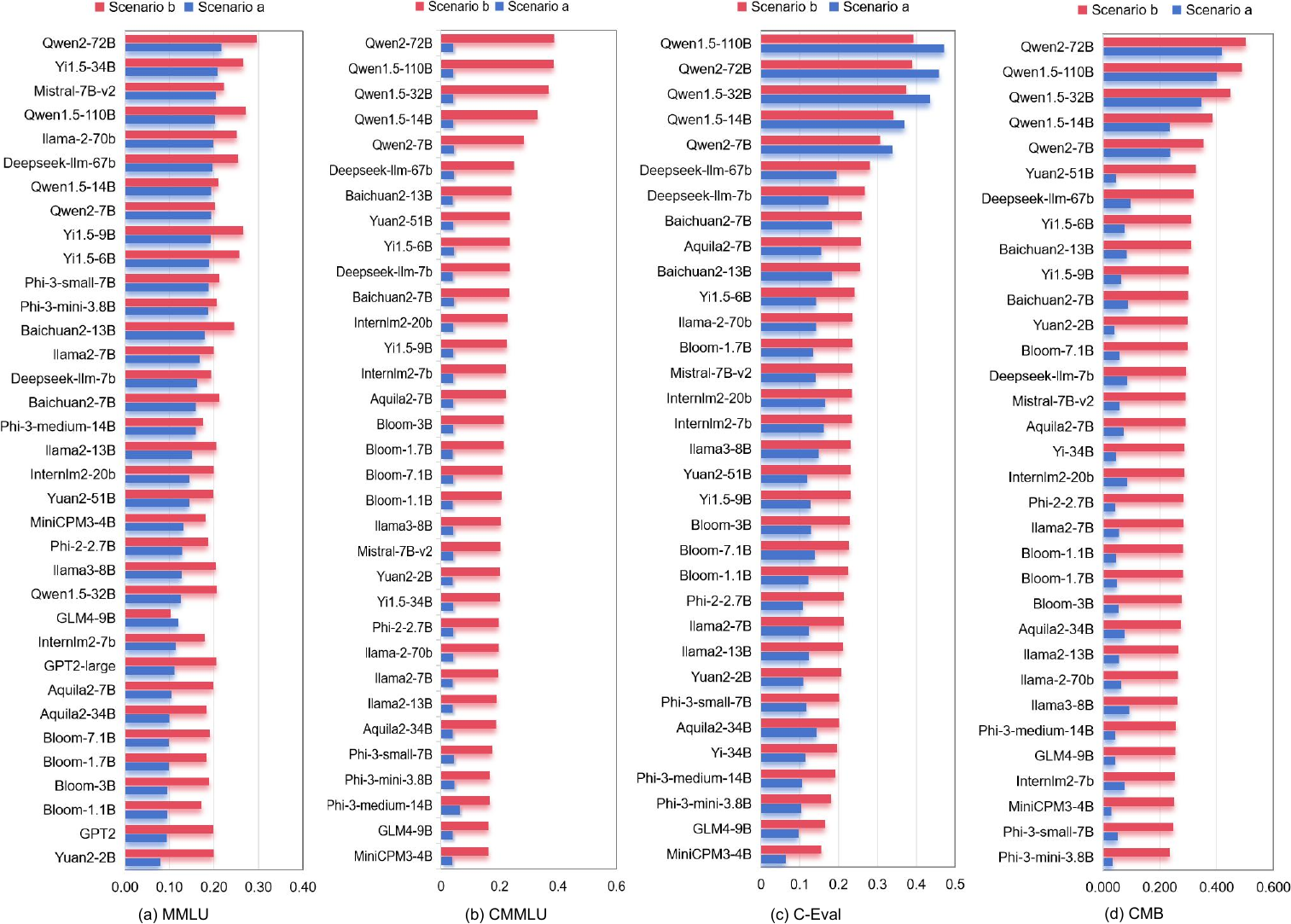} 
	\caption{Benchmark leakage leaderboard in LLMs.}
	\label{fig3}
\end{figure*}

\begin{figure*}[t]
	\centering
	\includegraphics[width=0.99\textwidth]{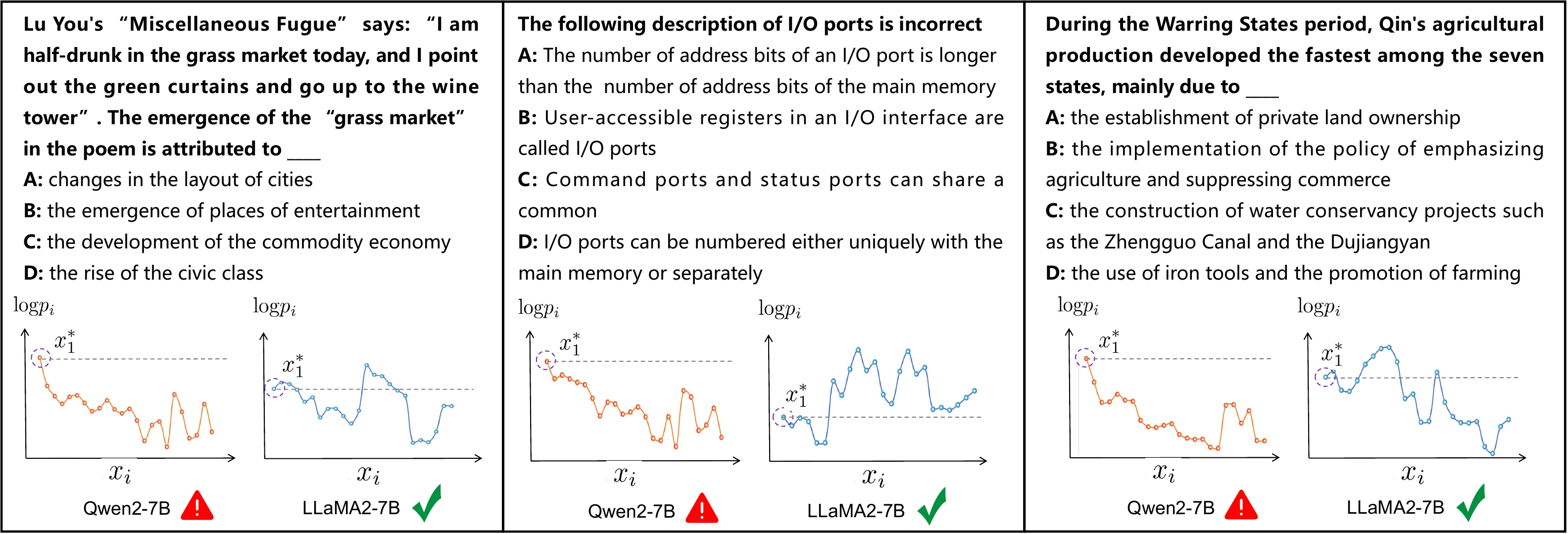} 
	\caption{Case analysis of Qwen2-7B and LLaMA2-7B on C-Eval under scenario \textit{a}.}
	\label{fig4}
\end{figure*}

\begin{figure}[t]
	\centering
\includegraphics[width=0.465\textwidth]{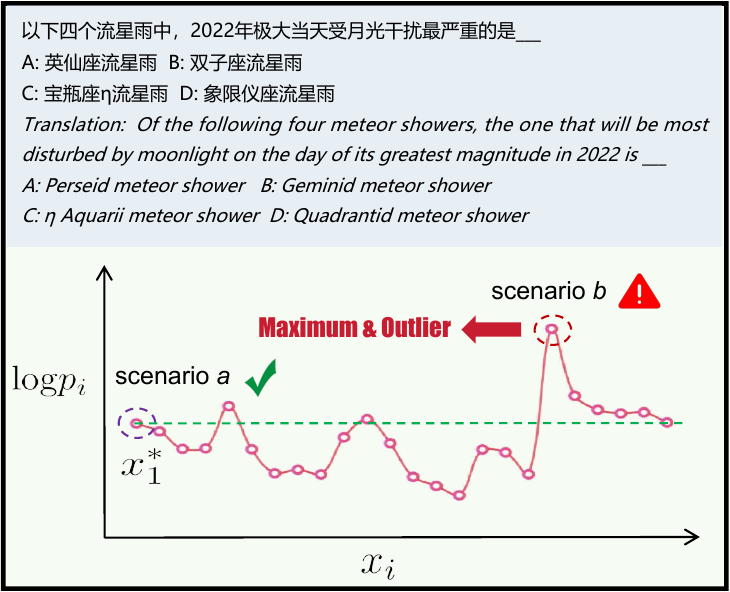} 
	\caption{Case analysis of Qwen2-7B under scenario \textit{b}.}
	\label{fig5}
\end{figure}
\section{Benchmark Leakage Leaderboard in LLMs}
The previous experiments demonstrate the effectiveness of our Algorithm \ref{alg1} and Algorithm \ref{alg2}, and next we will construct leaderboards for various benchmark leaks of LLMs.
We conduct comprehensive data leakage detection experiments on four mainstream benchmarks: MMLU \cite{hendrycks2021measuring}, CMMLU \cite{li2023cmmlu}, C-Eval \cite{huang2024c}, CMB \cite{wang2023cmb}. As shown in Figure 3, we tested almost all of the currently popular 35 LLMs \cite{team2023internlm,touvron2023llama,touvron2023llama2,young2024yi,BAAI,bai2023qwen,yang2023baichuan,bi2024deepseek,abdin2024phi}, and we give the percentage predicted to be data leakage for both scenarios \textit{a} and \textit{b}. The outlier threshold $\delta$ for our scenario \textit{b} is set to 0.2 on the three benchmark test sets, MMLU, CMMLU, and C-Eval; since there are five options for the data in the CMB benchmark, its outlier threshold $\delta$ is set to 0.25. And Ordered by the degree of leakage under scenario \textit{b}. The Benchmark leakage leaderboard in Figure 3 is sorted by the degree of leakage under the scenario \textit{b}. First of all, we find that there is not much gap between models on the MMLU benchmark, and the top five models in terms of data leakage risk are Qwen2-72B, Qwen1.5-110B, Yi-34B, Yi1.5-9B and Yi1.5-6B. Overall, the leakage of LLMs on the MMLU benchmark is a serious concern, and as MMLU is one of the most used and widely used benchmarks in the English language domain, the issue deserves our attention.

On the CMMLU benchmark, the leakage metrics shown on scenario \textit{a} are all very low, basically only about 0.04, which is basically in line with the expectation of 1/24 = 0.042 for normal conditions. We then found that the data leakage metrics detected under scenario \textit{b} were all significantly higher after detection using Algorithm \ref{alg2}, especially the Qwen family, which ranked the highest. We hypothesize that it is possible that the CMMLU benchmark shuffled the options after collecting the raw data or that the developers of LLM shuffled the pre-training data in a shuffling operation. On C-Eval, a Chinese comprehensive benchmark similar to CMMLU, the top five modeled data leakage risks are also all Qwen1.5-110B, Qwen2-72B, Qwen1.5-32B, Qwen1.5-14B and Qwen2-7B. On the Chinese Medicine Benchmark CMB, the top five LLMs in terms of data breach risk remain Qwen2-72B, Qwen1.5-110B, Qwen1.5-32B, Qwen1.5-14B and Qwen2-7B. In particular, the Qwen family of LLMs leads off the cliff, with Algorithm \ref{alg1} scoring much higher than the other models. In terms of data leakage values, the Qwen family LLMs are almost ten times larger than other LLMs. Algorithm \ref{alg1} detects that 42\% of the test data of the CMB benchmark on Qwen2-72B is leaked.

Overall, GLM4-9B and MiniCPM3-4B have the lowest risk of data leakage on all three benchmarks, MMLU, CMMLU, and C-Eval, and a low risk of data leakage on CMB. Qwen family LLMs have very high leakage risk on all 4 benchmarks, and we find that the larger the model the higher the leakage index, which might be due to the fact that larger models have more pre-training data and are more capable of learning and remembering the data more firmly. In addition to the Qwen family LLMs, the Yi family LLMs, DeepSeek family LLMs, and Baichuan family LLMs are also at slight risk of benchmark compromise. Mild benchmark leaks are hard to avoid, but we hope that researchers should avoid serious benchmark leaks when developing LLMs.
\section{Case Study}
As shown in Fig. 4, we select three examples from C-Eval in order to analyze the data leakage under scenario \textit{a} more intuitively. For example, in the first case the original data $x^*_1$ is “\textit{Lu You's “Miscellaneous Fugue” says: “I am half-drunk in the grass market today, and I point out the green curtains and go up to the wine tower”. The emergence of the “grass market” in the poem is attributed to \_\_\_\_A: changes in the layout of cities    B: the emergence of places of entertainment C: the development of the commodity economy   D: the rise of the civic class}” , and we shuffle the contents of the options to get 24 derived data $\mathcal{X} = \{x_1,x_2,... ,x_{n!}\}$. We then compute all possible shuffling combinations based on Qwen2-7B and LLaMA2-7B, respectively, to obtain two sets of (n!) logarithmic probabilities $\mathcal{P}_{Qwen} = \{\mathrm{log}p_1,\mathrm{log}p_2,\ldots,\mathrm{log}p_{n!}\}$ and $ \mathcal{P}_{LLaMA} = \{\mathrm{log}p_1, \mathrm{log}p_2, \ldots, \mathrm{log}p_{n!}\}$. A dot-line plot based on these two sets of log probabilities is shown in Figure 4. The logarithmic probability of the original sequential data $x^*_1$ on the Qwen2-7B model is the largest, larger than the logarithmic probability of any of the other 23 sequences, which suggests that this data is at risk of leakage on Qwen2-7B. On the right LLaMA2-7B's are the normal plots, when the contents of the options are shuffled, some of the log probabilities become smaller and some larger, and original sequential data $x^*_1$ is not the largest, which suggests that there is no data leakage from LLaMA under scenario \textit{a}.

A particular example of Qwen2-7B is shown in Figure 5, where the log probability of the original sequence $x^*_1$ is not the largest, and the entry is detected by Algorithm 1 as not leaking under scenario \textit{a}. However, our detection results using Algorithm \ref{alg2} detects that this piece of data is a leakage risk because the 19th derivation sequence has the highest log probability and is judged to be an outlier. The question and option for that piece of data is “Of the following four meteor showers, the one that will be most disturbed by moonlight on the day of its greatest magnitude in 2022 is \_\_\_  A: Perseid meteor shower   B: Geminid meteor shower C: $\eta$ Aquarii meteor shower  D: Quadrantid meteor shower”, and theoretically for the LLM being tested, the content of the shuffled option should not have a maximum log probability of being a significant outlier.
This case illustrates that our Algorithm \ref{alg2} is also effective in detecting data leakage for the case where the option content is shuffled. Primarily the intent of the leaderboard is to promote a fairer assessment of the community's LLMs, not to expose a particular model.

\section{Conclusion}
This work has highlighted the severity of benchmark data leakage in Large Language Models (LLMs) and introduced an innovative detection method capable of identifying leakages under various scenarios, including when the order of multiple-choice options may have been shuffled. We validate the effectiveness of the approach based on two LLMs design experiments and evaluate the data leakage risk of 35 open-source LLMs on four mainstream benchmark sets, present a benchmark leakage leaderboard among LLMs, and in particular find that the Qwen family of LLMs shows a high risk in several benchmarks. This work emphasizes the need for developers and researchers to be vigilant in ensuring the integrity and fairness of LLM assessments. We call for continued community effort to address this issue, improve our detection techniques, and uphold the robustness of benchmark assessments in the field of artificial intelligence. This paper serves as a stepping stone towards establishing more reliable and trustworthy standards in the evaluation of LLMs and advancing the field of artificial intelligence with confidence and integrity.
Currently our method is limited to detecting data in multiple choice format, in the future we will try to extend our method to other formats.
\section*{Acknowledgements}
This work was supported by Open Research Fund of Key Laboratory of Intelligent Education Technology and Application of Zhejiang Province, Zhejiang Normal University (Grant No. jykf21002w), National Natural Science Foundation of China (62376262), GuangDong Basic and Applied Basic Research Foundation (2023A1515110718 and 2024A1515012003), China Postdoctoral Science Foundation (2024M753398), Postdoctoral Fellowship Program of CPSF (GZC20232873).

\bibliography{aaai25}

\end{document}